\documentclass[conference]{IEEEtran}

\usepackage{cite}
\usepackage{amsmath,amssymb,amsfonts}
\usepackage{graphicx}
\usepackage{textcomp}
\usepackage{xcolor}
\usepackage{booktabs}
\usepackage{multirow}
\usepackage{url}
\usepackage{balance}
\usepackage{microtype}
\usepackage{xspace}
\usepackage{tabularx}
\usepackage{array}
\graphicspath{{figures/}}

\def\BibTeX{{\rm B\kern-.05em{\sc i\kern-.025em b}\kern-.08em
    T\kern-.1667em\lower.7ex\hbox{E}\kern-.125emX}}

\newcommand{\rmr}{RMR-P\xspace}
\newcommand{\mapfifty}{mAP50\xspace}
\newcommand{\mapfull}{mAP50--95\xspace}

\begin{document}

\title{RMR-P: Road Metadata-Aware Restoration for Pavement Inspection}

\author{
\IEEEauthorblockN{
Amir Ghorbani$^{\dagger,\ddagger}$,
Amirali K. Gostar$^{\dagger}$,
WeiQin Chuah$^{\dagger}$,
Vahid Ghorbani$^{\dagger}$,
Aidan Blair$^{\dagger}$,\\Reza Hoseinnezhad$^{\dagger}$,
Alireza Bab-Hadiashar$^{\dagger}$
}
\IEEEauthorblockA{
$^{\dagger}$School of Engineering, RMIT University, Melbourne, VIC 3000, Australia\\
$^{\ddagger}$Department of Infrastructure Engineering, The University of Melbourne, Melbourne, VIC 3052, Australia
}
}

\maketitle

\begin{abstract}
Road-surface images captured by vehicle-mounted cameras are often degraded by motion blur, defocus, poor illumination, and noise due to vehicle motion, camera limitations, and varying environmental conditions. These degradations can obscure thin cracks and pothole boundaries that are critical for accurate road-defect detection. This paper presents \rmr, a restoration network designed to recover defect-relevant information from degraded road images. It estimates degradation characteristics from the input image and can optionally incorporate external degradation parameters to guide restoration.
To evaluate whether the recovered information improves downstream detection, a clean-trained YOLO11s detector is applied to degraded and restored images without further modification. Experiments on the IVCNZ and PCM datasets, with known synthetic degradation parameters provided as conditioning information, demonstrate that \rmr achieves the highest mAP50 in seven of eight held-out degradation conditions, including improvements from 0.140 to 0.427 under IVCNZ motion blur and from 0.060 to 0.233 under PCM defocus. Moreover, our ablation studies show that preserving fine pavement details contributes most to defect-detection improvement, while degradation conditioning and task-guided training offer complementary benefits.

\end{abstract}


\section{Introduction}

Camera-based systems are increasingly being deployed across smart cities for a wide range of sensing, monitoring, and perception applications \cite{mien2022traffic,dang2024pedestrian,phuong2024vehicle,ishtiaq2022roadconstraints,ghorbani2025real}.
  Pavement inspection using camera-based systems provides a scalable approach for road maintenance planning and infrastructure monitoring. However, the reliability of vision-based defect detection depends strongly on image quality during acquisition. In practical road inspection scenarios, motion or defocus blur, poor illumination, sensor noise, and compression artifacts can reduce the contrast of subtle pavement structures, making thin cracks and pothole boundaries difficult to distinguish from surrounding road textures. Although recent road-damage datasets and deep learning detectors have significantly advanced automated inspection \cite{maeda2018road,arya2024rdd2022,chitale2020pothole,giordani2026pcm,ghorbani2026robust}, most detectors operate directly on captured images and do not explicitly recover visual evidence degraded during image acquisition.

Image restoration has achieved substantial progress through efficient network architectural design using convolutional or transformers layers \cite{nah2017deep,zamir2022restormer,chen2022nafnet}. However, existing restoration methods are generally optimized for visual fidelity, such as pixel-level similarity or perceptual quality. For road inspection, these objectives alone may not be sufficient. For example, an image that appears visually cleaner can still be unsuitable for detection if restoration removes or smooths out narrow cracks or weak defect boundaries. Therefore, restoration for inspection applications should focus not only on improving image appearance but also on preserving visual evidence that supports downstream defect recognition. To this end, recognition-aware restoration methods try to preserve task-relevant information \cite{wang2022togethernet,yang2023vrdir}. Furthermore, inertial-assisted deblurring studies have shown that acquisition-related information can reduce restoration ambiguity \cite{joshi2010image,hu2016image,mustaniemi2019gyroscope}.

One potential direction is to make restoration adaptive to the characteristics of the degradation affecting each image. This is because different degradations, such as blur, defocus, noise, and illumination changes, may remove different types of visual information and thus require different restoration strategies. Therefore, restoration can benefit from explicitly representing degradation characteristics and using them to guide the recovery process. This motivates a degradation-aware restoration approach that considers both the underlying degradation and the requirements of downstream defect detection.

In this work, we propose \rmr, a compact restoration network designed to recover defect-relevant information from degraded road images. Unlike conventional restoration approaches that optimize primarily for image quality, \rmr explicitly models degradation characteristics and uses them to guide restoration for road defects. The network estimates degradation information from the input image and can optionally incorporate available degradation parameters when provided. A detail-preserving pathway selectively restores fine pavement structures while limiting excessive enhancement, helping maintain the visual evidence required for defect detection.

The proposed framework is evaluated using a frozen road-defect detector to measure whether restoration improves downstream detection performance. Experiments are conducted on the IVCNZ \cite{chitale2020pothole} and PCM \cite{giordani2026pcm} road-damage datasets under controlled degradation conditions. The results demonstrate that degradation-aware restoration can substantially improve defect detection accuracy compared with directly applying detectors to degraded images.

\section{Method}

\subsection{Controlled Degradation Conditioning}
\label{sec:controlled_conditioning}

Let $I_d\in[0,1]^{3\times H\times W}$ denote a degraded road image and
$I_c$ its clean training target. RMR-P first estimates the corruption directly
from the image:$\hat{z}=g_{\phi}(I_d), \hat{z}\in[0,1]^8.$ The eight coordinates describe horizontal motion, vertical motion, motion magnitude or irregular vibration, defocus, noise, low illumination, compression, and overall severity. On controlled training pairs,
$\hat{z}$ is supervised against the known generator state $z^{*}$. This gives
the image encoder an interpretable target rather than allowing it to learn an unconstrained latent representation. Each controlled image is also accompanied by a small degradation-parameter
record $m$. This record is written by the program that creates the corrupted
image. Depending on the degradation, it can contain motion-blur length and
direction, defocus strength, illumination gain, gamma, noise level, or
compression quality. A fixed normalization and encoding function converts
this record into $z_m=g_m(m), \qquad z_m\in[0,1]^8.$ For example, one executed IVCNZ motion-blur sample was generated with
\begin{equation}
m=
\left\{
\text{blur length}=19\ \text{pixels},\quad
\text{blur angle}=29.50^{\circ}
\right\},
\end{equation}
while its defocus, noise, low-light, and compression fields were zero. The
blur magnitude is first normalized as $19/25=0.76$. Under the axial
directional encoding used in the implementation, this record produces
approximately
\begin{equation}
z_m=
[0.576,\;0.706,\;0.760,\;0,\;0,\;0,\;0,\;0.760].
\label{eq:numerical_metadata_example}
\end{equation}
Thus, the model is informed that the image contains strong directional motion
blur, but it is not given the clean image, defect annotations, detector
predictions, or desired restoration.

The image estimate and supplied degradation code are combined using a
coordinate-wise reliability gate:
\begin{equation}
z=\rho\odot z_m+(1-\rho)\odot\hat{z},
\label{eq:reliability_fusion}
\end{equation}
where $\rho\in[0,1]^8$ is predicted from the agreement between $z_m$ and
$\hat{z}$ together with a record-availability flag. A large value of
$\rho_j$ gives more weight to the supplied record for coordinate $j$, whereas
a small value makes the network rely primarily on the image estimate.

Metadata dropout and small perturbations of $z_m$ are applied during training
so that the network does not assume that every supplied value is exact
\cite{ghorbani2023data}. When the record is missing or inconsistent with the
image, the model can reduce its influence and rely more strongly on
$\hat{z}$. In a physical deployment, $m$ would instead be formed from available camera
and vehicle measurements, for example exposure time, gain, vehicle speed,
gyroscope, accelerometer, and pose. 


\subsection{Conditioned Restoration and Bounded Detail}
The restoration backbone is a compact three-scale encoder--decoder constructed from depthwise-separable residual blocks \cite{howard2017mobilenets}. The fused state modulates each block through feature-wise linear modulation (FiLM) \cite{perez2018film}:
\begin{equation}
\operatorname{FiLM}(x,z)=(1+\gamma(z))\odot x+\beta(z).
\end{equation}
Four label-free evidence maps---gradient magnitude, local contrast, darkness, and colour saturation---guide a lightweight task-evidence attention branch toward weak pavement structures without requiring detector inference at deployment.

The decoder first predicts a base restoration $I_b$. To reduce oversmoothing, fixed local mean filters form a multi-scale high-frequency residual bank. A small branch converts the residual bank and evidence maps into a learned detail signal $D(I_d)$ and a bounded spatial gate $G_d\in[0,1]^{H\times W}$. Defining $\eta_d(z)=\eta_{\max}\rho_d(z)$, the final output is
\begin{equation}
I_r=\operatorname{clip}\!\left(I_b+\eta_d(z)G_d\odot D(I_d),0,1\right),
\end{equation}
where $\eta_{\max}$ caps the complete update and $\rho_d(z)$ reduces detail transfer when the estimated degradation makes copied texture unreliable. The path can therefore retain a thin crack or pothole rim without becoming an unrestricted sharpening filter.

\subsection{Task-Aware Training and Optional Measurement}
A dataset-specific YOLO11s detector is trained on clean images and frozen before restoration training. The principal reconstruction term combines pixel, edge, frequency, and weak-road-evidence consistency. A low-weight detector-feature term compares intermediate frozen-detector features of $I_r$ and $I_c$. A Hutchinson-style Jacobian penalty discourages small image changes from producing disproportionately large detector-feature changes:
\begin{equation}
\mathcal{L}_{J}=\mathbb{E}_{v}\left\|\nabla_{I_r}
\left\langle\Phi(I_r),v\right\rangle\right\|_2^2.
\end{equation}
The optimized objective is summarized as
\begin{equation}
\mathcal{L}=\mathcal{L}_{\mathrm{base}}+\lambda_z\mathcal{L}_{\mathrm{code}}
+\lambda_{\mathrm{det}}\mathcal{L}_{\mathrm{det}}
+\lambda_J\mathcal{L}_{J}+\mathcal{L}_{\mathrm{safe}},
\end{equation}
where $\mathcal{L}_{\mathrm{safe}}$ denotes low-weight feature anchoring, evidence non-regression, and detail-containment safeguards. The fidelity term remains dominant, and the detector weights are never updated.

\begin{figure*}[!t]
\centering
\includegraphics[width=0.75\textwidth]{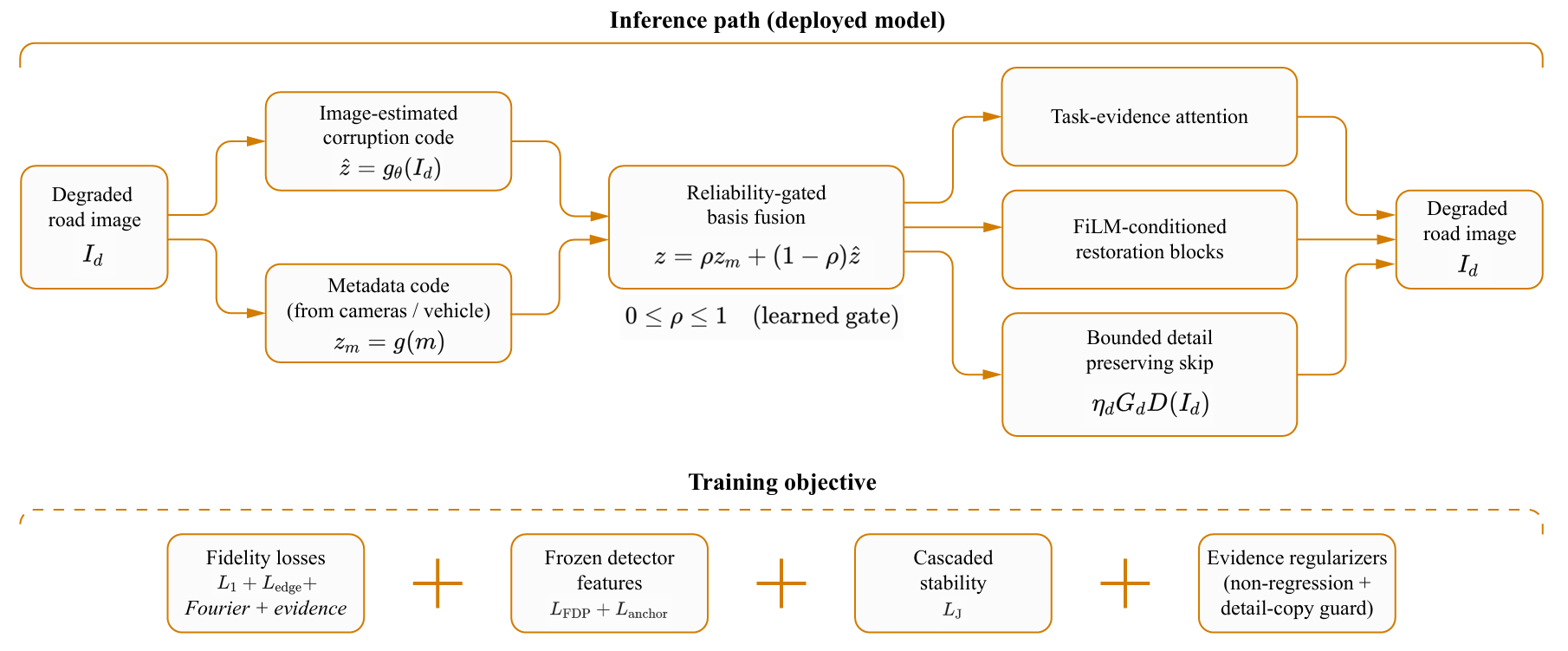}
\caption{Overview of \rmr. }
\label{fig:architecture}
\end{figure*}

Figure~\ref{fig:architecture} summarizes the core architecture components. 

\section{Experimental Setup}
\subsection{Datasets and Splits}
Table~\ref{tab:data} gives the exact image-level partitions. The IVCNZ pothole data contain 1,243 ordered frames \cite{chitale2020pothole}. PCM contains 2,009 images at $640\times360$ resolution with pothole, crack, and maintenance-hole annotations \cite{giordani2026pcm}; polygons are converted to boxes for detector evaluation. All clean images are split before any corruption is generated, so no degraded derivative of a test image enters training or validation. Labels are unchanged across clean, degraded, and restored versions.

\begin{table}[!t]
\centering
\caption{Controlled dataset partitions.}
\label{tab:data}
\scriptsize
\setlength{\tabcolsep}{3.5pt}
\begin{tabular}{lccc}
\toprule
Dataset & Train & Validation & Test \\
\midrule
IVCNZ images & 870 & 186 & 187 \\
IVCNZ boxes  & 2,699 & 693 & 684 \\
PCM images   & 1,406 & 301 & 302 \\
PCM boxes    & 3,321 & 703 & 712 \\
\bottomrule
\end{tabular}
\end{table}

\subsection{Synthetic Degradations and Conditioning Records}
Four road-relevant corruption families are generated after splitting: (i) parameterized line-spread motion blur, (ii) isotropic defocus blur, (iii) multiplicative low-light reduction with noise, and (iv) motion blur combined with low light and noise. Three normalized severity levels, 0.30, 0.60, and 0.90, are used. The recorded state includes corruption family, severity, blur direction and length, defocus strength, illumination reduction, and noise level when applicable. Motion length is represented relative to a 25-pixel reference and direction is encoded by horizontal and vertical components. These saved generator values form $m$ and $z^{*}$; they do not contain labels, clean pixels, detector outputs, or real sensor measurements. 

\begin{table*}[!t]
\centering
\caption{Information supplied to each method at inference in the controlled
IVCNZ and PCM experiments. The external conditions are selected or constructed
from the benchmark degradation record; they do not contain clean images,
defect annotations, or detector outputs.}
\label{inference_information}

\footnotesize
\setlength{\tabcolsep}{5pt}
\renewcommand{\arraystretch}{1.20}

\begin{tabularx}{\textwidth}{
    >{\raggedright\arraybackslash}p{2.45cm}
    >{\centering\arraybackslash}p{1.25cm}
    >{\raggedright\arraybackslash}p{3.65cm}
    >{\raggedright\arraybackslash}X
}
\toprule
\textbf{Method} &
\textbf{Image} &
\textbf{External information} &
\textbf{Use at inference} \\
\midrule

Clean reference
&
$I_c$
&
None
&
Passed directly to the frozen detector as an upper-bound reference. 
\\

Degraded input
&
$I_d$
&
None
&
Passed directly to the frozen detector without restoration.
\\

DFPIR
&
$I_d$
&
Benchmark-selected text prompt
&
The scenario family selects a fixed prompt, such as ``Blur degradation with
motion blur'' or ``Lowlight degradation.'' DFPIR encodes this prompt using
CLIP and uses the resulting embedding to guide feature perturbation. It does
not receive blur length, direction, or other continuous parameters.
\\

NAFNet-road
&
$I_d$
&
None
&
Predicts the restoration solely from the degraded image, without an external
prompt, degradation code, or task label.
\\

DeMoE-A
&
$I_d$
&
None
&
Recognizes the degradation internally from $I_d$ and dynamically weights or
selects its decoder experts.
\\

DeMoE-S
&
$I_d$
&
Benchmark-provided degradation-family label
&
The scenario family manually selects one supported expert, such as synthetic
global motion, defocus, or low light, instead of using DeMoE's automatic
router. No continuous degradation parameters are supplied.
\\

RMR-P image-only
&
$I_d$
&
None
&
The capture state is estimated internally as
$\hat{z}=g_{\phi}(I_d)$ and used to condition restoration.
\\

\textbf{RMR-P}
&
$I_d$
&
Eight-dimensional capture record $z_m$
&
The external record is reliability-gated with the image estimated code to
form the fused representation . 
\\

\bottomrule
\end{tabularx}

\vspace{3pt}
\begin{minipage}{0.97\textwidth}
\footnotesize
\textit{Notes:}
The RMR-P record is
$z_m=[m_x,m_y,v,d,n,\ell,c,s]$, representing horizontal motion,
vertical motion, irregular vibration, defocus, noise, low illumination,
compression, and overall severity; inactive coordinates are zero.


\end{minipage}
\end{table*}

\subsection{Detector, Restorers, and Selection Protocol}
A separate YOLO11s detector \cite{redmon2016yolo,ultralytics2024yolo11} is trained for each dataset on its clean training partition for 80 epochs using $640$-pixel letterboxing, batch size 8, and seed 2026. The checkpoint with the best clean-validation \mapfull is frozen. Every clean, degraded, baseline-restored, and \rmr image is then evaluated with the same detector using confidence 0.001 and class-aware non-maximum suppression at IoU 0.70.

DFPIR \cite{tian2025dfpir} and DeMoE \cite{feijoo2026demoe}\textbf{ use their released checkpoints }and receive no road-data optimization. DeMoE-auto routes from the image, whereas DeMoE-scenario is supplied the correct corruption-family label and is therefore an informed-routing comparison. NAFNet-road \cite{chen2022nafnet} is trained from random initialization on the same paired road split for 12 epochs with AdamW, batch size 6, $192$-pixel crops, learning rate $2\times10^{-4}$, and seed 2026.

\rmr is trained for 30 global epochs using AdamW, automatic mixed precision, batch size 1, $128$-pixel crops, seed 2026, a base learning rate of $10^{-6}$, and a tenfold multiplier for newly introduced heads. Each epoch is saved; the checkpoint maximizing mean frozen-detector validation \mapfifty across the four corruption families is selected. Both reported controlled runs select global epoch 28, and the test sets are evaluated once after selection. The primary endpoint is frozen-detector \mapfifty; the clean-test score is reported as an upper reference, not as a restoration result.

\section{Results and Discussion}
\subsection{Complete Controlled Comparison}
Table~\ref{tab:fullresults} reports every tested scenario and image source. \rmr gives the highest non-clean \mapfifty in seven of eight conditions. On IVCNZ, it raises motion-blur \mapfifty from 0.140 to 0.427, defocus from 0.154 to 0.372, and mixed degradation from 0.280 to 0.427. On PCM, the largest gains occur for defocus (0.060 to 0.233), low light (0.326 to 0.435), and mixed degradation (0.061 to 0.139). The retained exception is PCM motion blur, where DeMoE-scenario reaches 0.320 and \rmr reaches 0.312. The full table therefore supports strong but not universal superiority. See Table \ref{inference_information} for more details on supplied information to each model during experiments.

\begin{table*}[!t]
\centering
\caption{Complete held-out frozen-YOLO11s detection comparison. Values are \mapfifty. DeMoE-A/S denote automatic and known-scenario routing; NAFNet-R denotes road-trained NAFNet. Bold indicates the best non-clean image source in each row.}
\label{tab:fullresults}
\scriptsize
\setlength{\tabcolsep}{4.1pt}
\begin{tabular}{llccccccc}
\toprule
Dataset & Degradation & Clean ref. & Degraded & DFPIR & DeMoE-A & DeMoE-S & NAFNet-R & \rmr \\
\midrule
\multirow{4}{*}{IVCNZ}
& Motion blur & 0.652 & 0.140 & 0.374 & 0.371 & 0.382 & 0.190 & \textbf{0.427} \\
& Defocus & 0.652 & 0.154 & 0.211 & 0.248 & 0.254 & 0.357 & \textbf{0.372} \\
& Low light & 0.652 & 0.517 & 0.475 & 0.510 & 0.510 & 0.505 & \textbf{0.523} \\
& Motion + low light & 0.652 & 0.280 & 0.307 & 0.343 & 0.351 & 0.275 & \textbf{0.427} \\
\midrule
\multirow{4}{*}{PCM}
& Motion blur & 0.541 & 0.203 & 0.294 & 0.319 & \textbf{0.320} & 0.194 & 0.312 \\
& Defocus & 0.541 & 0.060 & 0.091 & 0.136 & 0.152 & 0.174 & \textbf{0.233} \\
& Low light & 0.541 & 0.326 & 0.352 & 0.317 & 0.263 & 0.364 & \textbf{0.435} \\
& Motion + low light & 0.541 & 0.061 & 0.059 & 0.082 & 0.080 & 0.083 & \textbf{0.139} \\
\bottomrule
\end{tabular}
\end{table*}

\begin{figure*}[!t]
\centering
\includegraphics[width=0.8\textwidth]{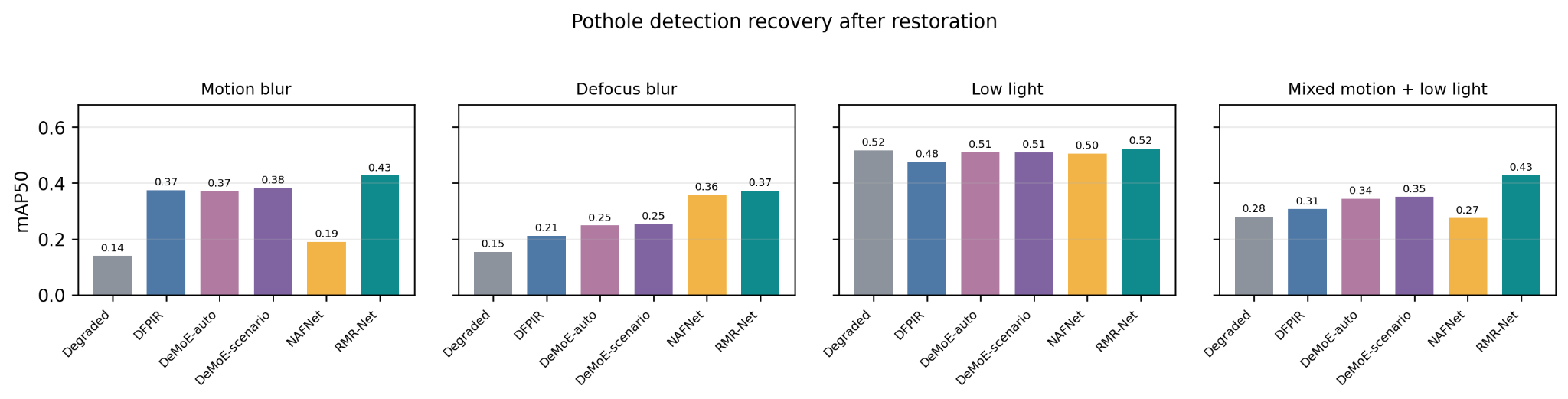}\\[-1mm]
\includegraphics[width=0.8\textwidth]{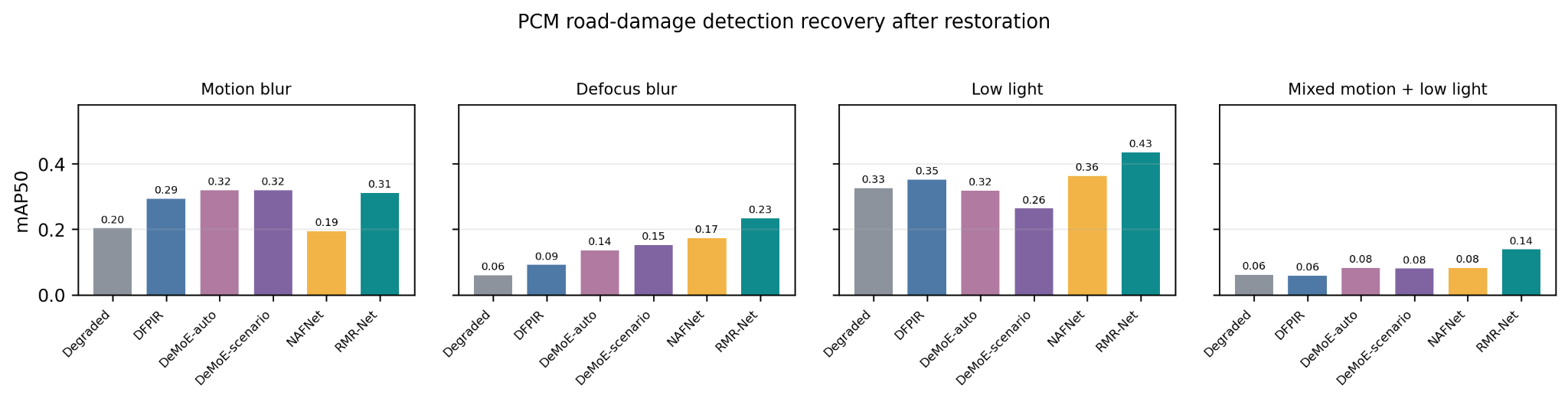}
\caption{Scenario-wise frozen-detector recovery. Top: IVCNZ pothole benchmark. Bottom: PCM pothole/crack/manhole benchmark. }
\label{fig:recoveryplots}
\end{figure*}

Figure~\ref{fig:recoveryplots} visualizes the same complete comparison scenario by scenario. The IVCNZ pothole plot confirms the large motion, defocus, and mixed-degradation recovery, while the PCM plot makes the motion-blur exception and stronger defocus, low-light, and mixed results explicit. The labels above the bars are displayed to two decimals for readability; the authoritative three-decimal values are those in Table~\ref{tab:fullresults}. In the plots, ``NAFNet'' denotes NAFNet-road.

The same frozen detector is used throughout, so the gains reflect changes in image evidence rather than detector retraining. DFPIR, DeMoE, and NAFNet remain important controls because stronger visual smoothing does not always translate into stronger detector recovery.

\subsection{Component Ablation}
Table~\ref{tab:ablation} gives a compact sequential ablation on PCM defocus. All trained rows use the same split, seed, eight-epoch budget, and validation-\mapfifty selection. This shorter diagnostic protocol is separate from the 30-epoch checkpoint. We report the sequence intentionally, even though it is non-monotonic: modules that improve fidelity or stability need not independently increase test AP.

Supervised image-state learning and FiLM increase \mapfifty from 0.233 to 0.254. Adding controlled-parameter fusion gives 0.251, while the matched image-only control reaches 0.252; hence this specific run does not justify a broad claim that auxiliary parameters are always beneficial. The bounded detail path produces the largest local gain, from 0.240 to 0.264. Detector-aware loss and the final stability component then trade a small amount of AP for detector-feature and output safeguards. The last component includes the Jacobian term together with tightly coupled low-weight anchoring and non-regression terms.

\begin{table}[ht]
\centering
\caption{Sequential PCM-defocus ablation under a shared eight-epoch diagnostic protocol.}
\label{tab:ablation}
\scriptsize
\setlength{\tabcolsep}{3.0pt}
\begin{tabular}{lcccc}
\toprule
Configuration & \mapfifty & \mapfull & PSNR & SSIM \\
\midrule
Base restorer & 0.233 & 0.081 & 26.61 & 0.777 \\
+ image state and FiLM & 0.254 & 0.086 & 26.63 & 0.780 \\
+ controlled-parameter fusion & 0.251 & 0.086 & 26.66 & 0.781 \\
+ task-evidence attention & 0.240 & 0.085 & 26.81 & 0.783 \\
+ bounded detail path & \textbf{0.264} & \textbf{0.095} & \textbf{26.91} & \textbf{0.789} \\
+ detector-aware loss & 0.259 & 0.091 & 26.85 & 0.788 \\
+ stability package incl. $\mathcal{L}_J$ & 0.262 & 0.089 & 26.80 & 0.787 \\
\midrule
Matched image-only control & 0.252 & 0.088 & 26.77 & 0.784 \\
\bottomrule
\end{tabular}
\end{table}

\subsection{Qualitative Evidence and Boundary Output}
\begin{figure*}[!th]
\centering
\includegraphics[width=0.8\textwidth]{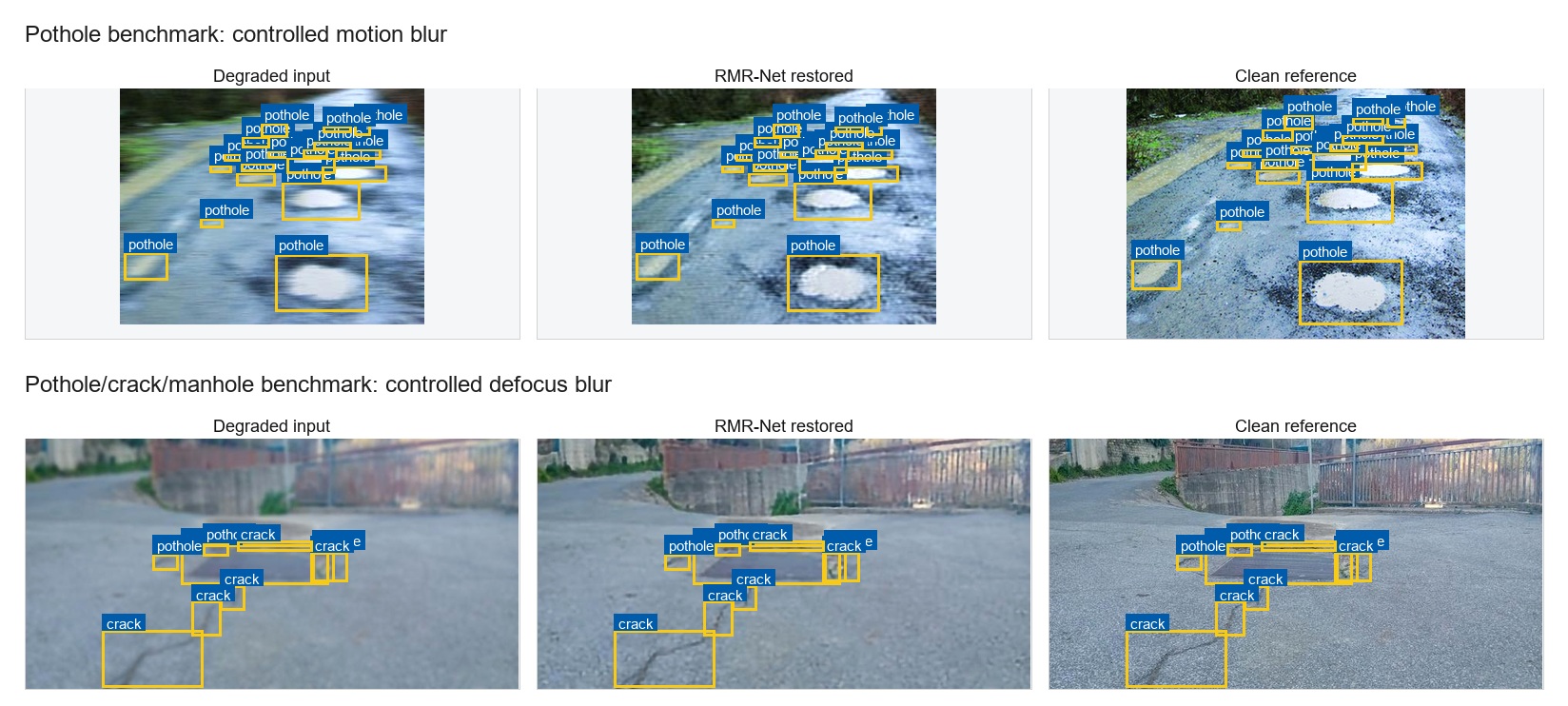}\\[1mm]
\includegraphics[width=0.8\textwidth]{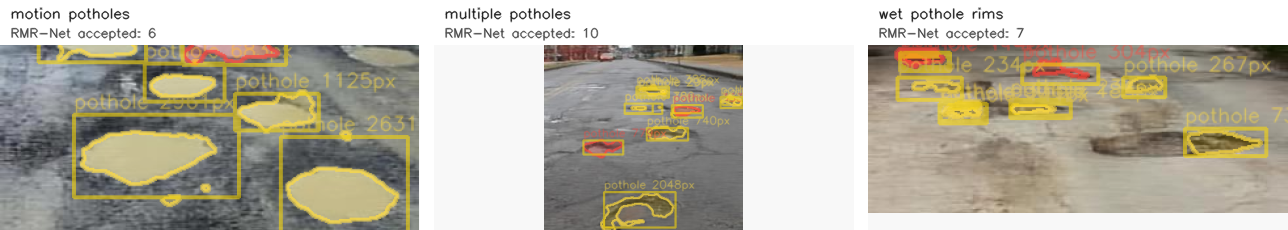}
\caption{Qualitative restoration. Top: controlled examples comparing degraded input, direct \rmr restoration, and clean reference. Bottom: representative guarded active-contour refinements initialized from accepted detector boxes. Boundary refinement is post-detection and does not modify the \mapfifty results.}
\label{fig:qualitative}
\end{figure*}

Figure~\ref{fig:qualitative} complements the aggregate metrics. The upper panel compares degraded input, direct \rmr output, and the corresponding clean reference for controlled motion blur and defocus. Yellow rectangles identify annotated defect regions; the restoration visibly recovers pothole rims, local road texture, and crack contrast toward the clean reference. These examples are illustrative, whereas the full test-set detector effects are reported in Table~\ref{tab:fullresults} and Fig.~\ref{fig:recoveryplots}.

The lower panel shows representative detector-guided contour outputs after restoration. The counts printed in the three panels are example-specific accepted contours, not additional test-set AP values. This optional measurement layer converts selected boxes into approximate boundaries for area, perimeter, and compactness estimates and remains strictly downstream of the frozen-detector evaluation. A qualitative examination of the contour results indicates that the method is working as expected and preserves the defect boundaries.

\subsection{Scope, Complexity, and Limitations}
The reported restorer has approximately 1.34 million parameters and requires about 114~ms per $640\times360$ frame on an NVIDIA RTX~3050 GPU. The controlled image-level splits provide reproducible within-source degradation evidence, but retrospective audits found a small number of duplicate or temporally adjacent frames across partitions. The results should therefore not be interpreted as field generalization.

Furthermore, the detector’s accuracy is closely linked to the model’s performance. Using a more accurate frozen detector during training is therefore expected to improve the model’s overall performance. Lastly, the experiments validate generator-aligned degradation conditioning, not real telemetry-conditioned restoration. Speed, angular rate, vibration, exposure, camera pose, GPS, and other synchronized vehicle measurements were not used to produce Table~\ref{tab:fullresults}. A future study should collect these signals jointly with natural blur and route-disjoint road imagery, quantify metadata reliability, and compare image-only, raw-telemetry, and physically derived priors under the same detector protocol.

\section{Conclusion}
This paper presented a light, task-focused model for controlled road-image restoration. The network estimates degradation evidence, fuses it with existing corruption context(if available), conditions an efficient restorer, and preserves local pavement detail through a bounded path. The main additional information that the proposed model receives compared to benchmarks is eight numbers ($z_m$)regarding degradation context which in practice is estimated from metadata and/or degrade image ( Table \ref{inference_information}). A clean-trained frozen detector provides low-weight training guidance and unchanged evaluation. Complete results across two datasets, four degradation families, and all baselines show the highest \mapfifty in seven of eight conditions, while the ablation identifies the bounded detail path as the strongest local contributor. Qualitative restoration and boundary examples illustrate how the recovered evidence can support inspection output beyond a visually cleaner frame and box detection. A promising future work is to train a separate model for estimating degradation context which will make the existing model a very promising tool for practical road inspection applications.

\section*{Acknowledgment}
This work is supported by the outcomes of the project \emph{NRSAGP-RD1-A50---Enhancing Road Safety Through Fleet-Based Vehicle Sensor Technology: A Strategy for Monitoring Road Conditions in Australia}, funded by the Australian Government. The authors would like to acknowledge Pouya Bagheri, Infographic Designer, for his valuable
contributions to the creation of graphics used in this paper.

\balance
\bibliographystyle{IEEEtran}
\bibliography{references}

\end{document}